# Uniform vs. Lognormal Kinematics in Robots: Perceptual Preferences for Robotic Movements

Jose J. Quintana, Miguel A. Ferrer, Moises Diaz, Jose J. Feo, Adam Wolniakowski and Konstantsin Miatliuk

**Abstract**

Collaborative robots or cobots interact with humans in a common work environment. In cobots, one under-investigated but important issue is related to their movement and how it is perceived by humans. This paper tries to analyze whether humans prefer a robot moving in a human or in a robotic fashion. To this end, the present work lays out what differentiates the movement performed by an industrial robotic arm from that performed by a human one. The main difference lies in the fact that the robotic movement has a trapezoidal speed profile, while for the human arm, the speed profile is bell-shaped and during complex movements, it can be considered as a sum of superimposed bell-shaped movements. Based on the lognormality principle, a procedure was developed for a robotic arm to perform human-like movements. Both speed profiles were implemented in two industrial robots, namely, an ABB IRB 120 and a Universal Robot UR3. Three tests were used to study the subjects' preference when seeing both movements and another analyzed the same when interacting with the robot by touching its ends with their fingers.

## 1. Introduction

The natural interaction between humans and robots is one of the demands produced because of the irruption of Industry 4.0. Previously, most robots were confined to industrial environments and were mostly related to other machines. In these environments, there was no need for the robot to relate to humans, usually for health and safety reasons. However, human–robot interaction (HRI) [1–3] has meant an industrial advance conquering large sectors of society and trying to make this interaction allow a "cordial" relationship between both actors [4].

One of the strategies pursued in HRI is trying to make robots appear human, that is, they have a trunk, head and limbs. These strategies also include providing robots with a humanoid voice or facial gestures. The goal is also to render robots as autonomous as possible, a goal that has been aided by tremendous advances in artificial intelligence intervening to support developments related to planning tasks or other more complex actions. On the other hand, there has also been progress in their relationship with humans using sounds, speech, digital screens, ubiquitous systems, and many other features.

Among the different factors involved in HRI, perhaps the least mediatic, but no less important is related to the humanoid movement of robots [5]. This movement is perceived unconsciously by humans and is not limited just to the robot's arms, but also extends to other movements such as those of the robot's head, eye blinking, mouth movement, hands, etc. [6,7].

In industrial environments, cobots are increasingly growing. The main difference between them and industrial robots is that the latter are designed to work in environments with no humans and the cobots are intended to be able to interact with them. For this reason, robots have safety measures so that humans do not cross their working area and if they do, they usually stop completely. In contrast, cobots, such as the URx series of Universal Robots [5], are designed to work with humans and can be configured so that they have zones in which they work without limitations, others in which they can interact with humans and others prohibited.

Due to the interaction with humans, the movement of cobots becomes more important. One of the ways to improve these interactions is to humanize the movement of robots when they perform any action [8]. This makes it easier for a human to predict the intention or unconsciously understand the robot's movement [9]. One of the main differences between human and robot movements lies in their kinematics. A robotic arm, for example, is programmed, as a rule, with a trapezoidal speed profile. That means its speed is constant in the path, except at the beginning and at the end, when its acceleration is constant. For its part, a human's speed profile when executing a movement is bell-shaped. Consequently, when the movement is long and complex, the shape of the speed profile is an ordered sum of bells, one for each individual movement

One theory widely used to describe this bell-shaped speed profile is the kinematic theory of rapid human movements, along with its Sigma-Lognormal model [10–14]. Thistheory explains most of the basic phenomena underlying human motor control and allowsto study various factors involved in fine motor skills. The lognormal model allows the analysis of human movement and to generate movements in a human fashion [15]. Basedon this model, the present work describes how to generate human movements and howto implement them in a robotic arm. The model was implemented in the ABB IRB120 industrial robot and verified with sensors.

In the previous paragraph, many references are made to the dynamic aspects of humanmovement. These dynamic aspects were not analyzed in this paper because they are notnecessary for the robot to make the movements and also because there is no access to thedynamic algorithms of the robot.

To check whether humans prefer robotic or human movement when interacting with robots, a series of tests were performed. Three tests analyzed the preference of the subjects when they see both movements: one looked at a point in a display, the other with the ABB IRB 120 robot and the last one using a Universal Robot UR3. A fourth test analyzed the same, but when the subject interacted with the robot by touching its end with his finger.

This work is laid out as follows: The first section describes the human arm andits equivalence with the robotic one. In the second section, the robotic arm is modeled mathematically. In the third, human movement is analyzed and the robot is programmedto perform human movements. In the fourth, several tests are carried out to verify the human perception of human and robotic movements. Finally, the discussion of the data iscarried out and conclusions are presented.

## 2. Human Arm

The human arm is made up of bones, joints and muscles. All these structures must work in coordination to produce movements such as those involved in writing. Bones and cartilage are rigid parts that shape and support the entire structure of the arm. The bone of the upper arm is the humerus and its central part is formed by the ulna and radius and the final part consists of the bones of the hand. The shoulder joint connects the scapula to the humerus, while the elbow joint connects the humerus to the ulna and radius. These two bones connect with the carpal bones, forming the wrist. Finally, the last joints are in the hand and join the carpal bones, metacarpals and finger bones [16].

The joints of the arm are the connections between the rigid parts of the arm and allow its movement through the muscles. The shoulder joins the arm to the shoulder. It is the most flexible joint in the human body and allows free movement of the arm. Under normal conditions, the arm hangs vertically from this joint. The elbow is the central joint of the arm and has a range of action going from 0 to 180º. The carpals and metacarpals, together with the rotation of the wrist, allow the hand to orient itself in all directions. Finally, the joints of the fingers allow one to grasp almost any object.

*Human and Robotic Arm Equivalence*

Due to its great versatility, the human arm has served as the inspiration for most robotic arms. The robot analyzed in the present work is an industrial robot with 6 degrees of freedom, manufactured by ABB, model IRB120. It was chosen because it is a robot based on the PUMA robot (Programmable Universal Manipulation Arm) and has a structure similar to most actual industrial robots. Its appearance can be seen in Figure 1.

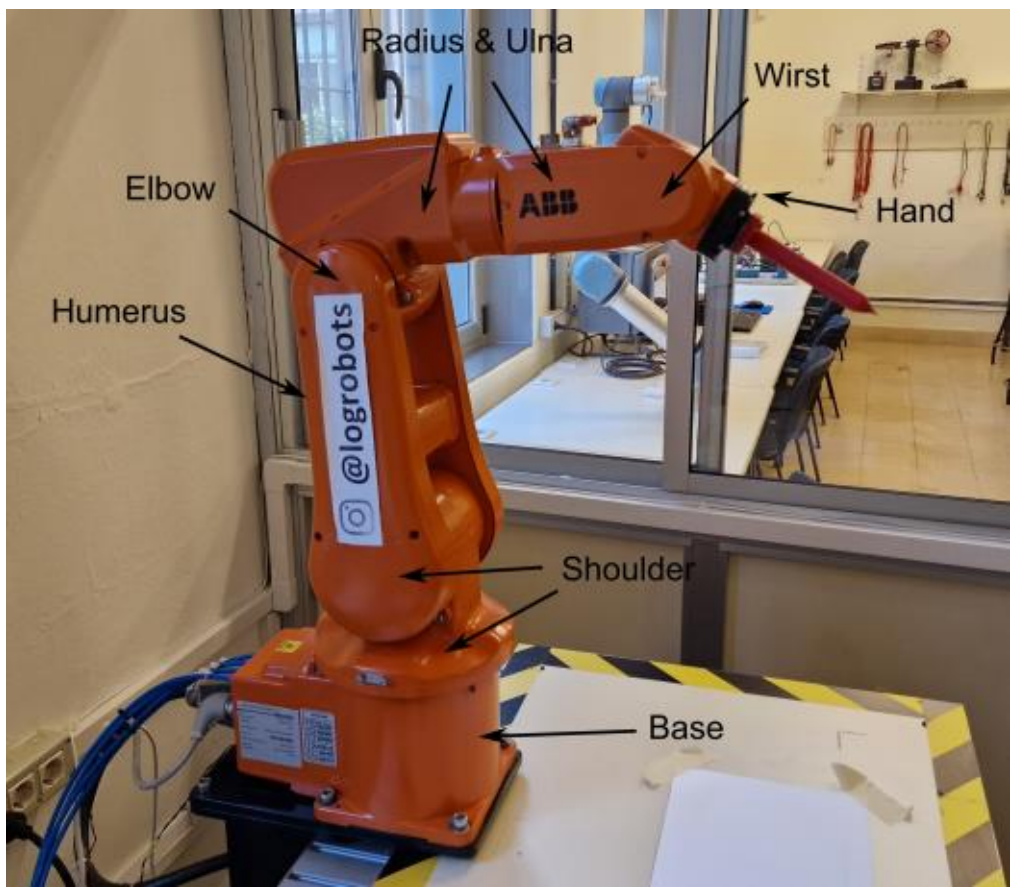

*Figure 1. Equivalence between robot and human arm*

The human arm can be considered as a redundant robot; that is, it has more degrees of freedom than are actually needed to orient and position an object in space. This means that a human arm provides theoretically infinite configurations to grasp a pencil in a certain position and with a certain orientation. This is not the case with industrial robotic arms, in which the number of possible configurations for holding a pencil at a certain position and orientation is limited. The human shoulder joint has its reflection on the robot in its first two joints. The first joint of the robot is a rotation of the first link related to the base and on a vertical axis and thesecond joint is a rotation of the second link related to the first. These two joints allow two movements of the second link with respect to the base, the equivalent of the humerus with respect to the scapula. In any case, the robotic arm has a degree of freedom less than the human shoulder that also allows a turn along the humerus. The third joint is the one that relates the second and third links and is the equivalent of the human elbow. The rotational movement that allows the wrist to rotate with respect to the elbow is obtained in the robot with the rotation of the fourth link with respect to the third. The rotation of the wrist with respect to the forearm is obtained by rotating the fifth link with respect to the fourth andthe rotation of the pencil is obtained by rotating the sixth link with respect to the fifth.

## 3. Robotic Arm Modelling

To give an overview of the modelling of the robotic arm, a brief review of the mathematical tools to define the pose of an object, that is, its position and orientation will be shown. Next, the equations that relate the pose of the tool located at the end of the robotto the position of its joints will be deduced.

### *3.1. Pose of an Object*

To spatially define the tool located at the end of the robot, it is necessary to know its pose. Meanwhile, for the position of a point in space, its definition requires a coordinate frame (CF). This position is given in Cartesian coordinates by the projections of the pointon the three coordinate axes. In the case of a robotic arm, this CF is usually located at the base of the robot.

To define the orientation of an object, two CFs are necessary, with one (the fixed CF)taken as a base and another (a mobile CF) rigidly attached to the object whose orientation is to be defined and both must be placed with the same origin of coordinates. The position and orientation of the object are independent magnitudes, meaning that modifying either one does not imply a modification of the other and that both can be calculated separately. The rotation matrix is one of the most used methods to define orientations in robotics.

The CF OXYZ is defined as fixed and the OUVW as mobile. The rotation matrix is a $3\times3$ matrix in which the first column represents the unit vector û, the second column the unit vector v̂ and the third the unit vector ŵ, all related to the fixed CF.

Figure 2 shows the mobile and fixed CFs and both placed with the same origin.

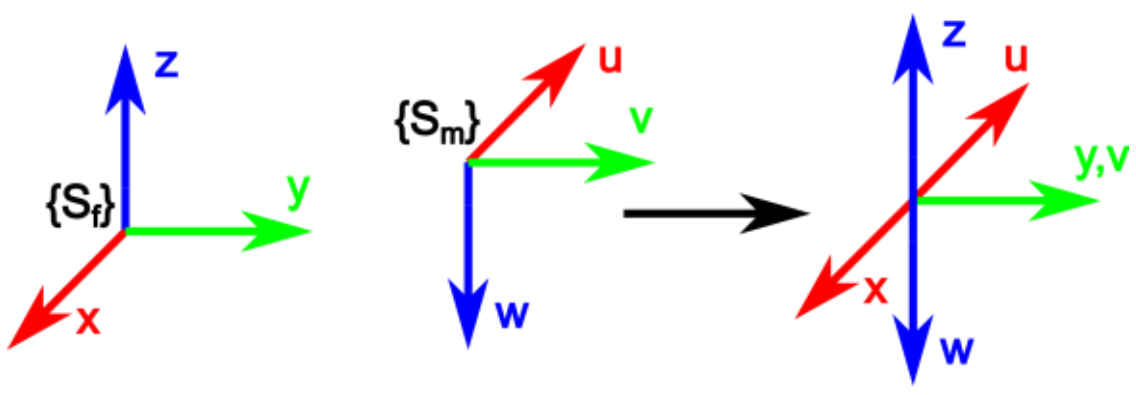

*Figure 2. Example of CFs to define the orientation of a mobile object.*

The rotation matrix that relates both CFs is given by,

$$R = \begin{pmatrix} n_x & o_x & a_x \\ n_y & o_y & a_y \\ n_y & o_z & a_z \end{pmatrix} = \begin{pmatrix} -1 & 0 & 0 \\ 0 & 1 & 0 \\ 0 & 0 & -1 \end{pmatrix} \tag{1}$$

It is noted that the first column indicates that the û vector is oriented in the negative direction of the x-axis (n = $(-1, 0, 0)$) , the second column indicates that the v̂ vector is in
the same direction as the y-axis and the third column indicates that the ŵ vector points in
the opposite direction to the z-axis (a = (0, 0, −1)).
If both mobile and fixed CFs have the same origin of coordinates, then if the coordinatesof a point in the mobile CF are known, they can be obtained in the fixed system as follows:

$$\begin{pmatrix} r_x \\ r_y \\ r_z \end{pmatrix} = R \begin{pmatrix} r_u \\ r_v \\ r_w \end{pmatrix} \tag{2}$$

As discussed above, the pose of an object is defined by the position and orientationof its CF with respect to the fixed CF. The mathematical tool used to define the pose of an object is the homogeneous transformation matrix, defined as:

$$T = \begin{pmatrix} R & p \\ f & w \end{pmatrix} = \begin{pmatrix} n_x & o_x & a_x & p_x \\ n_y & o_y & a_y & p_y \\ n_y & o_z & a_z & p_z \\ 0 & 0 & 0 & 1 \end{pmatrix} \tag{3}$$

where the matrix R of 3 3 is the rotation matrix described above, the vector p indicatesthe position of the mobile coordinate frame referring to the fixed CF, the vector f representsthe perspective vector, which is usually a null vector in robotics and finally, the scalar wrepresents the scaling factor, which always has a unit value in robotics.

As an example, the matrix that relates the mobile CF to the fixed one shown in Figure 3 is given by the homogeneous transformation matrix:

$$f_{T_m} \begin{pmatrix} 0 & 0 & -1 & L_1 \\ -1 & 0 & 0 & -L_2 \\ 0 & 1 & 0 & 0 \\ 0 & 0 & 0 & 1 \end{pmatrix} \tag{4}$$

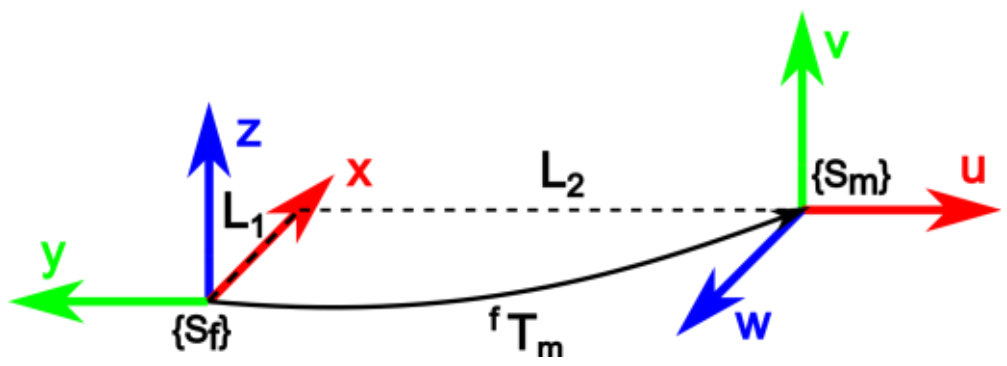

*Figure 3. Example of CFs to define the pose of a mobile object*

The nomenclature $^{f}T_{m}$ indicates the homogeneous transformation matrix that relates the pose of the mobile CF to the fixed one.

*3.1.1. Forward Kinematic*

The forward kinematics of a robot calculates the pose of the tool located at its end referred to the CF located at its base, as a function of its joints' position.
Forward kinematics is calculated by applying the procedure described by Denavit– Hartemberg (DH) [17] and which is widely used in robotics. In broad strokes, the procedure is as follows:
As shown in Figure 4, a CF is associated with the base of the robot $\{s_0\}$ ,one with
each joint in the corresponding link and one at the end of the tool $\{s_6\}$ . Therefore, in the six-degrees-of-freedom robot analyzed, there will be seven CFs. These coordinate frameswere located following the DH rules.

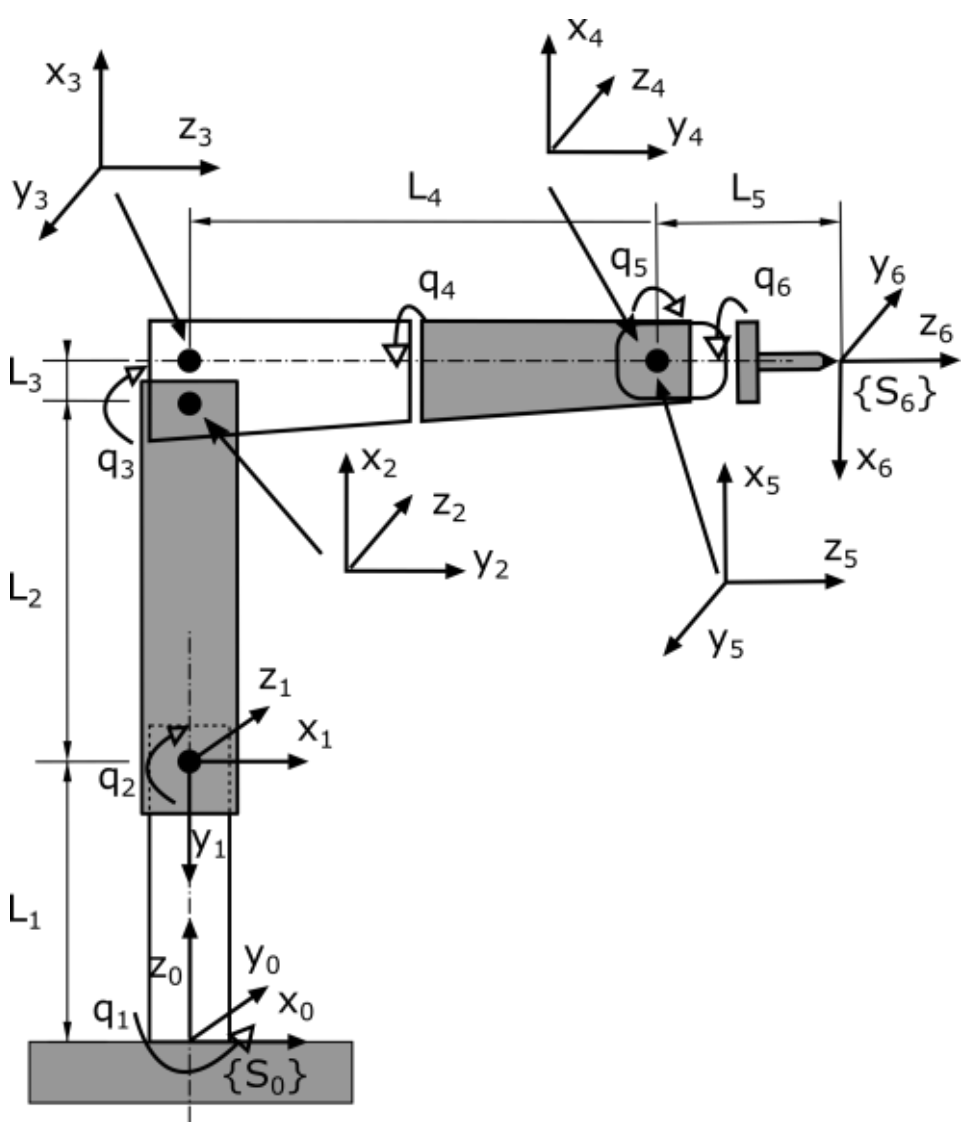

*Figure 4. Coordinate frames associated with the robot.*

These parameters are placed in a table in which each row represents a joint and the fourcolumns the respective values of each transformation. The following four transformations must be done on the $\{S_{k-1}\}$ to match the CF $\{S_k\}$:

1. Rotate an angle θ on the z-axis of the CF $\{S_{k-1}\}$ so that its x-axis is parallel to the x-axis of the CF $\{S_k\}$.

2. Displace the new CF $\{S_{k-1}\}$ to a given distance d on its z-axis so that its x-axis coincides with the x-axis of the CF $\{S_k\}$.

3. Displace the new CF $\{S_{k-1}\}$ to a given distance a on its x-axis so that the positions of both CFs coincide.

4. Rotate the new CF $\{S_{k-1}\}$ an angle α on the new x-axis so that the CFs match. With these parameters applied to the robot of Figure 4, Table 1 is obtained:

*Table1. Denavit–Hartemberg parameters.*

| Joint | $\theta$ | $d$ | $a$ | $\alpha$ |
|---|---|---|---|---|
| 1 | $q_1$ | $L_1$ | 0 | $-\pi/2$ |
| 2 | $q_2 - \pi/2$ | 0 | $L_2$ | 0 |
| 3 | $q_3$ | 0 | $L_3$ | $-\pi/2$ |
| 4 | $q_4$ | $L_4$ | 0 | $\pi/2$ |
| 5 | $q_5$ | 0 | 0 | $-\pi/2$ |
| 6 | $q_6 + \pi$ | $L_5$ | 0 | 0 |

Row $k$ of Table 1 has the information needed to calculate the homogeneous transformation matrix that relates the CF $\{S_k\}\{Sk\}$ taking the CF $\{S_{k-1}\}\{Sk-1\}$ as a reference and is given by:

$$^{k-1}T_k = \begin{pmatrix} c(\theta_k) & -c(\alpha_k)s(\theta_k) & s(\alpha_k)s(\theta_k) & a_k c(\theta_k) \\ s(\theta_k) & c(\alpha_k)c(\theta_k) & s(\alpha_k)c(\theta_k) & a_k c(\theta_k) \\ 0 & -s(\alpha_k) & c(\alpha_k) & d_k \\ 0 & 0 & 0 & 1 \end{pmatrix} \quad (5)$$

where c() is cos() and s() is sin(). Applying this equation to each of the rows in Table 1 and composing the matrices, we obtain the matrix that defines the pose of the tool located at the end of the robot with respect to CF of its base, depending on the angles of each of the joints.

$$^0T_6 = {}^0T_1\, {}^1T_2\, {}^2T_3\, {}^3T_4\, {}^4T_5\, {}^5T_6 \quad (6)$$

*3.1.2. Inverse Kinematic*

The goal of inverse kinematics is to deduce the position of each joint to obtain a given pose on the tool. In certain robots, these equations cannot be deduced in a closed fashion and they must therefore be calculated iteratively. In the case of the ABB IRB120 robot,the equations for each of the joints can be deduced and can be found at [18].

In a six-degrees-of-freedom robot such as the ABB IRB120, its tool can reach any pose that lies within its working area. Broadly speaking, this is due to the fact that the first three joints are mainly used to position the tool and the last three to orientate it. If the robot's tool only needs to describe a path regardless of the orientation, or it needs to set the values of some of the joints, then the method described in [18] is not valid and must be modified.

In this work, the last three joints are intended to be kept immobile and the position of the tool should only depend on the first three joints. That means that the robot will behave like a three-degrees-of-freedom system. To this end, the following tasks were performed in a Matlab script:

1. The variables of the last three joints were defined as global and kept constant:

$$q_{46} = [q_4, q_5, q_6] \quad (7)$$

2. The value of the point to calculate its inverse kinematic was inserted and declared asa global variable:

$$p = [p_x, p_y, p_z] \quad (8)$$

3. A function called ErrorFnct was created, with its parameter being a variable having the value of the first three joints and formatted as:

$$q_{13} = [q_1, q_2, q_3] \quad (9)$$

This function has as input parameters the positions of the first three joints. The position of the last three joints, $q_{46}$ and the point used to calculate the inverse kinematic p are passed to the function as global variables. This function, which uses forward kinematics, calculates and returns the error when applying a joint position. Algorithm 1 shows this function:

**Algorithm 1: Error function**

*global* $\mathbf{q_{46}}, \mathbf{p}$
*Function* : $ErrorFnct(\mathbf{q_{13}}) \leftarrow error$
$^0T_{pc} = FK(\mathbf{q_{13}}, \mathbf{q_{46}})$
$\mathbf{p_c} = {}^0T_{pc}(1:3, 4)$
$error = (p_x - p_{cx})^2 + (p_y - p_{cy})^2 + (p_z - p_{cz})^2$

4. The Matlab fminsearch function is used and introduces the function to be minimized and the seed value used to begin the iteration. The function returns the positions ofthe first three joints that minimize the error and has the following format:

$$q_{13c} = \text{fminsearch}(\text{'ErrorFnct'}, q_{130}) \quad (10)$$

5. By combining the positions of the first three joints with those of the last three, the po- sition of the robot is obtained:

$$q = [q_{13c}, q_{46}] \quad (11)$$

To calculate several consecutive points of a trajectory, the procedure above is repeated for each new point, but using the value obtained from the previous point as a seed.

## 4. Generating Movements with the Robot

One of the objectives of both current and future automation is to have robots interact with people. This carries many advantages since there are tasks in which the presence of a human facilitates automation and in the event of a breakdown or malfunction of the robotic cell, the fact that it can be accessed without the need to stop production increases productivity considerably. While a lot of effort has been invested in developing the sensors to apply in the cell and in their implementation in robot control software, one aspect that has not been given enough importance is the movement of robots. It has been shown that the speed profile of human movements is a flared one, meaning that the speed increases rapidly in the first stage of movement until it reaches a maximum and then decreases until it reaches its destination point. Conversely, the most common robotic movement usually has a profile characterized by a constant speed; at the beginning and at the end of the movement, the speed varies at a constant acceleration and during the movement, the speed remains constant. In this section, the procedure to implement these speed profiles in a robotic arm will be described.

### *4.1. Generating Human Movements*

There are many theories and models that have tried to describe human movement. Among these, the kinematic theory of rapid human movements and its Sigma-Lognormal model has been used to explain most of the basic phenomena reported in classical studies on human motor control and to study various factors involved in fine motor skills. The model herein proposes that a simple motion has a speed profile defined by a lognormal curve and that complex motions, such as signatures, can be decomposed into a sum of lognormals. In industrial robots, there are no standard movement commands based on a lognormal profile. To implement this profile on the robot, the lognormal curve and its parameters must be used:

$$v(t) = \Lambda(t; \mu, \sigma) \frac{1}{\sigma t\sqrt{2\pi}} \exp\left(-\frac{(\ln(t) - \mu)^2}{2\sigma^2}\right) \;\; \forall t > 0 \quad (12)$$

The lognormal distribution is a statistical distribution which implies that

$$\int_0^{\infty} \Lambda(t; \mu, \sigma) dt = 1 \quad \forall t > 0 \quad (13)$$

Therefore, if the speed has a lognormal distribution given by Equation (12), the final distance traveled will be the unit. If a scaling factor and a time lag are added to this distribution, the following equation results:

$$v(t) = \Lambda(t; t_0, \mu, \sigma) \frac{1}{\sigma(t - t_0)\sqrt{2\pi}} \exp\left(-\frac{(\ln(t - t_0) - \mu)^2}{2\sigma^2}\right) \;\; \forall t_0 > 0 \quad (14)$$

In this equation, t represents time, $t_0$ is the time in which the movement is activated, μ is the stroke time delay, σ is the stroke response time and D is a scaling factor. The parameters μ, σ, D and $t_0$ are calculated so that the speed profile has the desired temporal distribution. With Equation (14), the lognormal speed profile can be modeled, or in cases of having a human speed profile, characterize it with such a curve. To this end, the integral of the lognormal function given by its cumulative distribution function is used. This distribution is given by:

$$r(t) = \frac{1}{2}\left(1 + \text{erf} + \left(\frac{(\ln(t-t_0)-\mu)}{\sigma\sqrt{2}}\right)\right) \quad \forall t_0 > 0$$

$$r(t) = 0 \quad \forall t \leq t_0 \tag{15}$$

where the erf function is the statistical error function. We will assume that the trajectory that defines the end of the robot is a straight line from point $p_s$ to point $p_e$. Firstly, it will be necessary to define a lognormal curve that complies with the temporal parameters of speed and position, that is, to obtain the parameters of said curve. Considering that the cumulative distribution function starts from zero at time $t_0$ and reaches one at a time determined by the parameters μ and σ, the x-component of the trajectory is given by:

$$p_x(t) = p_{sx} + \frac{p_{ex}+p_{sx}}{2}\left(1 + \text{erf} + \left(\frac{(\ln(t-t_0)-\mu)}{\sigma\sqrt{2}}\right)\right) \quad \forall t_0 > 0$$

$$p_x(t) = p_{sx} \quad \forall t \leq t_0 \tag{16}$$

Operating analogously with the other components, the trajectories followed by $p_y$ and $p_z$ are obtained. Therefore, the procedure for the robot to follow a straight line with a lognormal speed distribution is as follows:

1. Define the starting $p_s$ and ending point $p_e$ of the trajectory and the time used to perform the movement $t_e$.
2. With the above parameters, define the parameters of a lognormal curve that meet the above requirements. It should be noted that there are infinite curves as a function of μ and of σ. Analyzing the behavior of both parameters, it was observed that the one with the greatest effect on the shape of the curve was σ and so a value of μ = ln(1) was set. Therefore, σ must be calculated to have the time of the movement as $t_e$. This is shown in Figure 5.

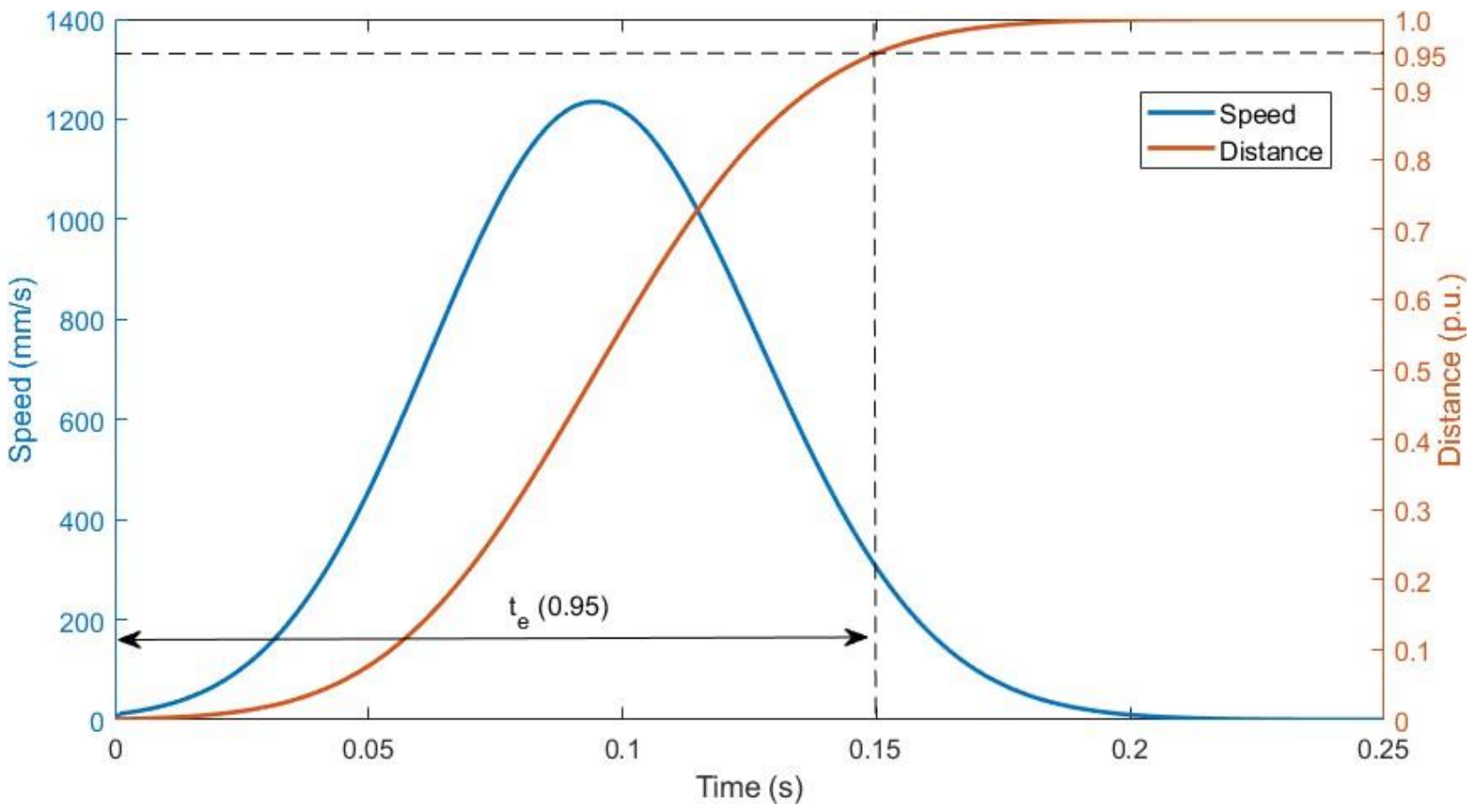

*Figure 5. Procedure for calculating the parameter σ*

The blue curve shows the lognormal speed profile, in which $t_0 = 0$ and $\mu = \ln(1)$. The red curve is the integral given by Equation (15), whose range is between zero, for t = 0 and one, for t = ∞. It is observed that although the time to reach 1 is theoretically infinite, this value is reached much earlier than that. In the figure, this value is set at time $t_e$ to 0.95. Knowing the values of $t_0$ and of μ, the value of σ is calculated with iterative methods. For clarity, an r value of 0.95 was used in the explanations, but in the tests, an r value of 0.99 was used.

3. With Equation (16) and knowing the values of the initial and end points, $t_0$, μ and σ the points of the trajectory and its corresponding time are calculated. Points will be obtained based on the selected sample rate.

4. For each point and setting the last three joints of the robot to constant values to ensure there are no reorientation effects, the inverse kinematics is applied, and its joint coordinates are obtained. These will be the coordinates that must be followed by each of the robot's joints in time.

5. The joint coordinates are applied to the robot, which must follow a rectilinear movement with a lognormal speed profile.

6. If the trajectory is composed of several straight lines, the last procedure is repeated with each of the lines, allowing a temporal overlap between them.

*4.2. Generating Trapezoidal Profiles*

The definition of a trapezoidal speed profile by the robot is not necessary since this is implemented in its standard commands. However, if we want to add some restrictions to this profile, as in this case, the speed profile must be generated.

The procedure for the robot to follow a straight line with a trapezoidal speed profile is as follows:

1) Define the starting $p_s$ and ending point $p_e$ of the trajectory and the time used to perform the movement.

2) The maximum acceleration at the tool is defined in the manufacturer datasheet and can be limited to a lower value. The speed profile and its corresponding displacement are shown in Figure 6, for a movement that starts and ends at rest. As seen in the figure, the movement initially proceeds at constant acceleration, until the desired speed is achieved and then the rest of the movement is performed with that constant speed and finally the robot returns to rest at constant acceleration. Using the speed in the module, the times at constant acceleration and at constant speed are calculated.

3) With these times, the speed profiles for the three components are generated; that is, the points of the trajectory and their corresponding times are calculated based on the selected sample rate.

4) With each point and with the last three joints of the robot set at constant values to prevent any reorientation effects, the inverse kinematics is applied, and its joint coordinates are obtained for each point. These will be the coordinates that each of the robot's joints must follow in time.

5) The joint coordinates are applied to the robot, which must carry out a rectilinear movement with a trapezoidal speed profile.

6) If the trajectory is composed of several straight lines, the last procedure is repeated with each of the lines, allowing a temporal overlap between them.

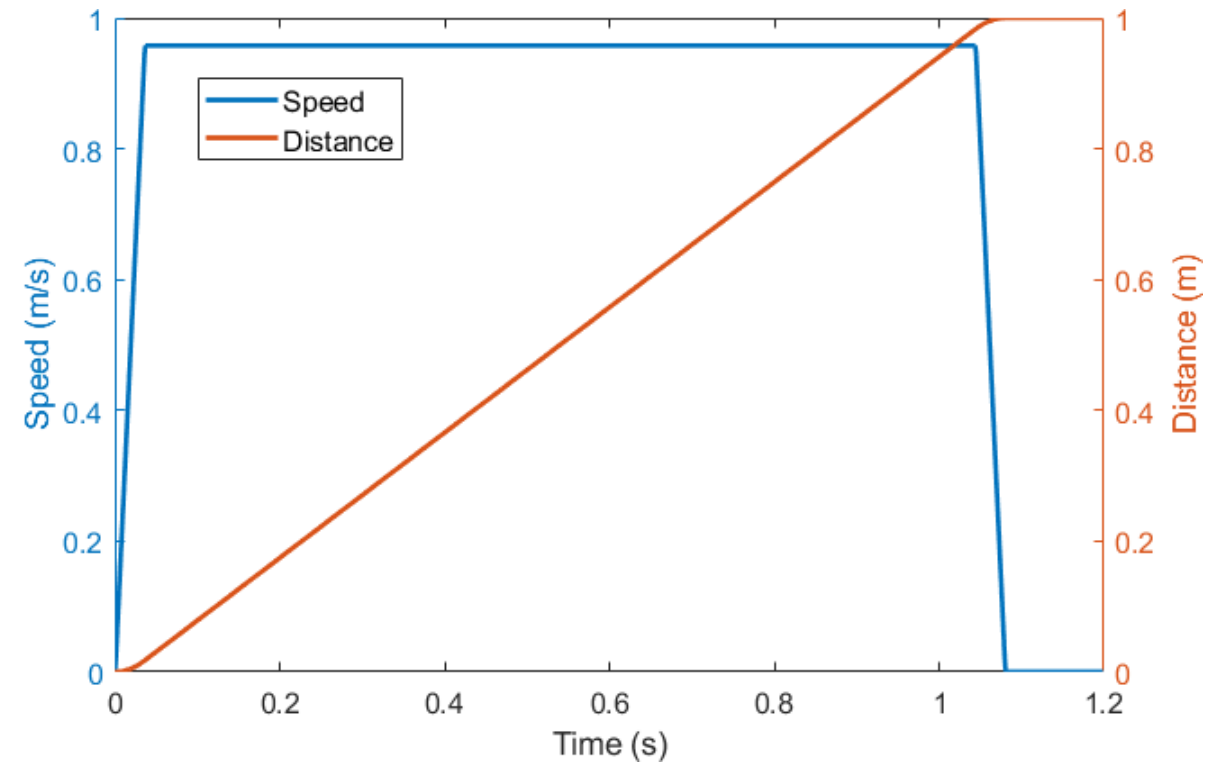

*Figure 6. Trapezoidal speed profile.*

*4.3. Applying the Speed Profiles to the Robot*

The proposed algorithms were implemented and tested in an ABB IRB120 robot. To test the speed profiles in it, several closed figures were defined in space, with their vertices joined by straight lines. The robot was programmed to describe each figure with a constant speed profile and with a lognormal one.

The trajectories programmed were defined with a sampling period of 24 ms. They were converted to Cartesian trajectories referring to the base of the robot. Then, using a Matlab script, the values of the first three joints were calculated considering that the last three joints were constant values. Therefore, each point of the trajectory had its corresponding point in articular coordinates. This trajectory was implemented in the robot and used to describe the desired movement.

The trajectories shown in Figure 7, with the two speed profiles, were programmed. To verify that the programmed movements were executed by the robot correctly, the sensor described in [19] was used. This sensor has an accuracy of a few millimeters and a sample rate of 200 samples per second.

As an example, trajectory B1 with a constant speed profile was programmed in the robot. The robot performed it and was recorded by the sensor. The trajectory captured by the sensor is shown in Figure 8.

Once it was verified that the robot performed the programmed figure, the next step was to check if the speed profile performed by the robot was the programmed one. To this end, a speed profile was programmed in the robot and the sensor was used to check whether the robot had executed it correctly. Figure 9 shows the trapezoidal speed profile implemented in the robot and the one recorded with the sensor. The similarity of the curves is obvious, but when assessing their similarity, the Signal-to-Noise Ratio (SNR) index, defined as follows, will be used:

$$\mathrm{SNR} = 10\log\left(\frac{\sum_{i=0}^{n} v_{p,i}^{2}}{\sum_{i=0}^{n}(v_{p,i} - v_{p,i})^{2}}\right) \tag{17}$$

where $v_{p,i}$ is the speed programmed at the instant i, $(v_{p,i} - v_{r,i})$ is the error made by the robot when executing the instruction and n is the number of samples in the file. The higher the SNR, the better the setting. The good fit in a visual way in the figure corresponds also to a high SNR value of 22.5.

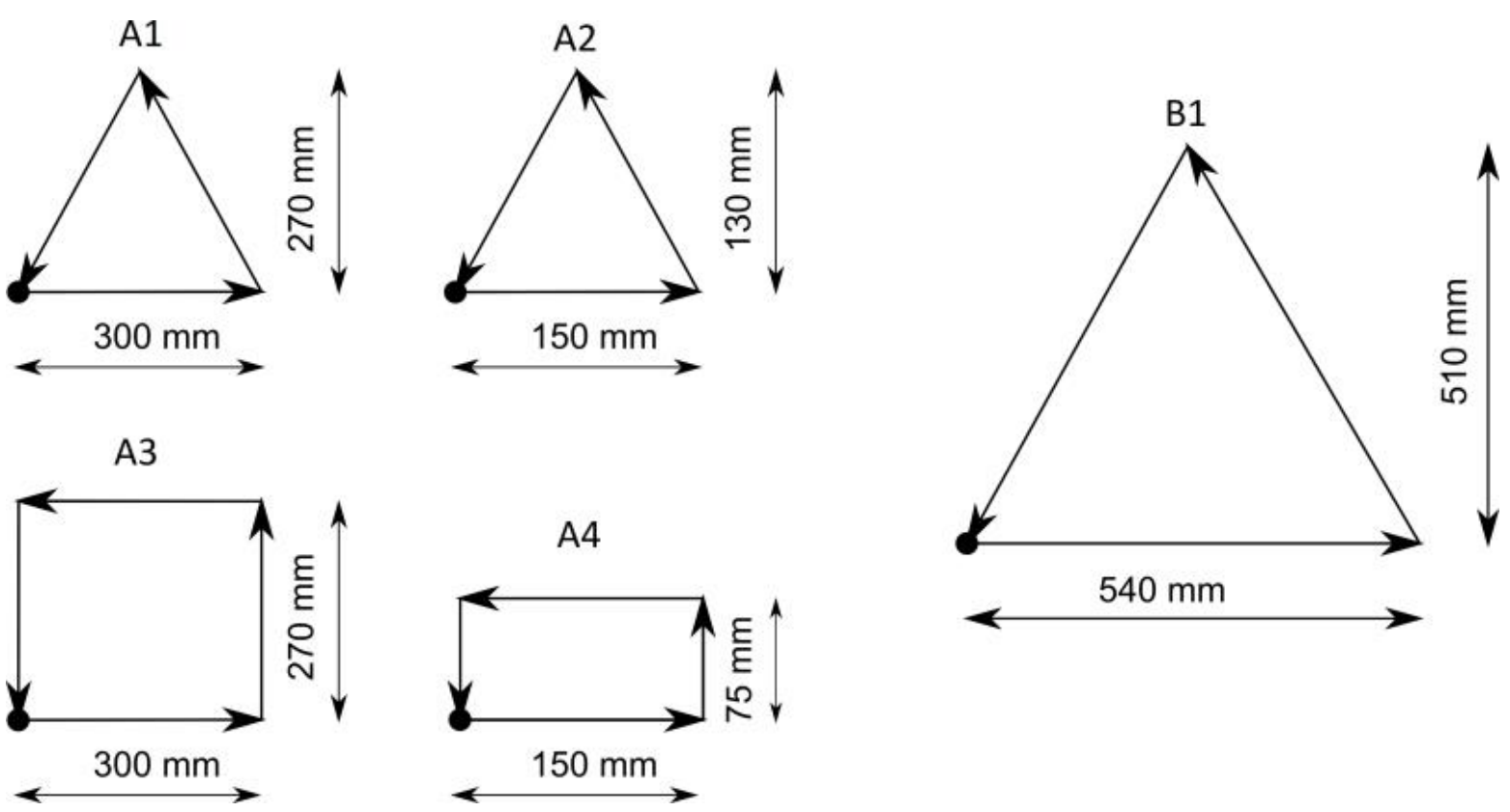

*Figure 7. Figures programmed in the robot.*

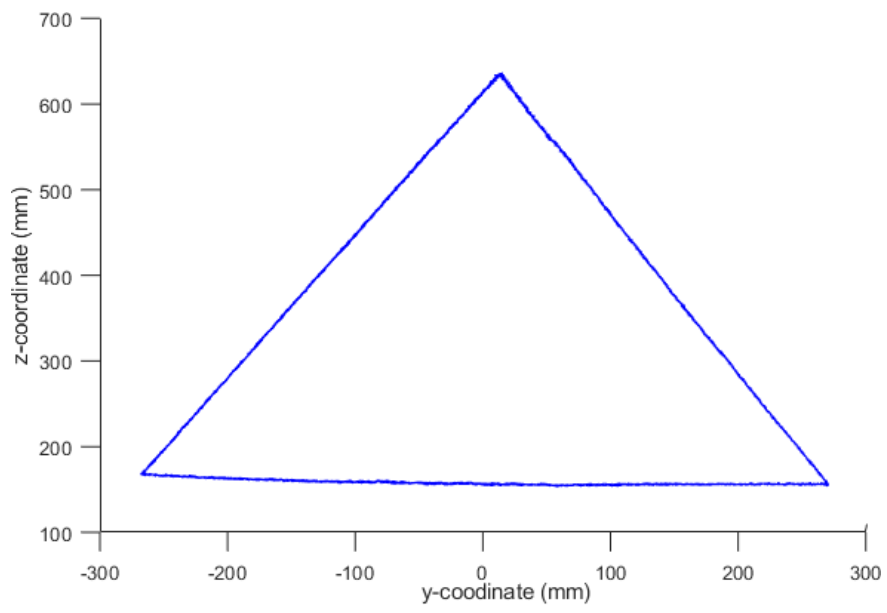

*Figure 8. Figure performed by the robot.*

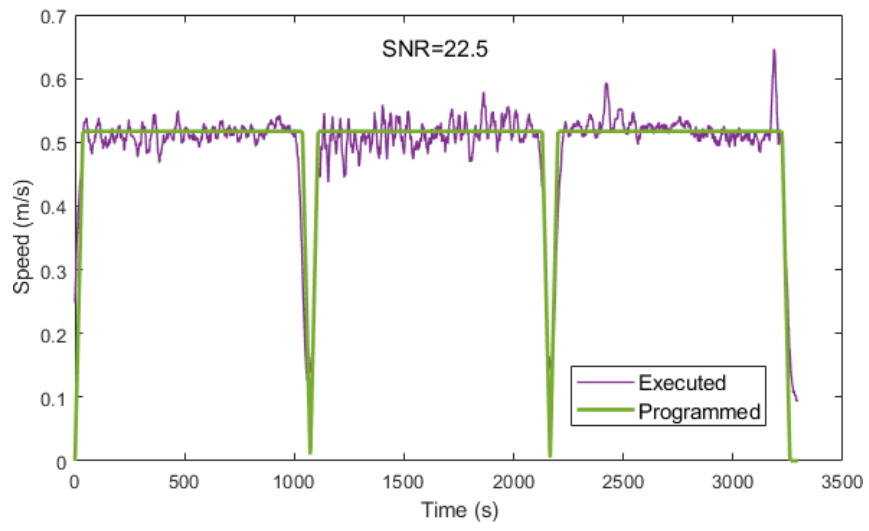

*Figure 9. Trapezoidal speed profile programmed and executed.*

Similarly, Figure 10 shows the programmed speed profile and the executed one whena lognormal speed profile is applied. A good fit can be observed both visually and with anSNR value of 23.6.

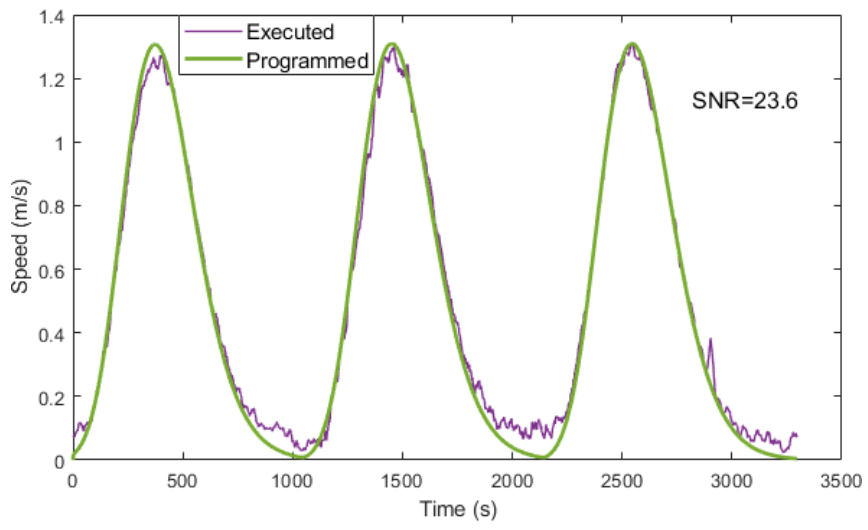

*Figure 10. Lognormal speed profile programmed and executed.*

It was thus proven that the ABB IRB120 robot was able to execute the programmedspeed profiles with great precision.

## 5. Tests Performed

As mentioned above, human movement has a lognormal speed profile, while robotic movement is usually performed at a constant speed. The tests carried out aimed to corroborate the following hypotheses:
Humans unconsciously detect robotic motion as an unnatural motion, leading to rejection.
Humans find it easier to intuit the trajectory of a robotic arm with a human speed profile than with a robotic one.
The interaction of a human with a robot is more friendly if the movement of the latteris human.
The first three tests were carried out using Google forms and in a virtual environment, as this was during the quarantine period and with Covid-19 restrictions in effect.

### *5.1. Test 1*

This test was carried out using Google forms (https://bit.ly/3n3OZj8, accessed on1 October 2022). It was based on the ABB IRB 120 robotic arm performing movements with uniform and lognormal speed profiles, as shown in Figure 11.

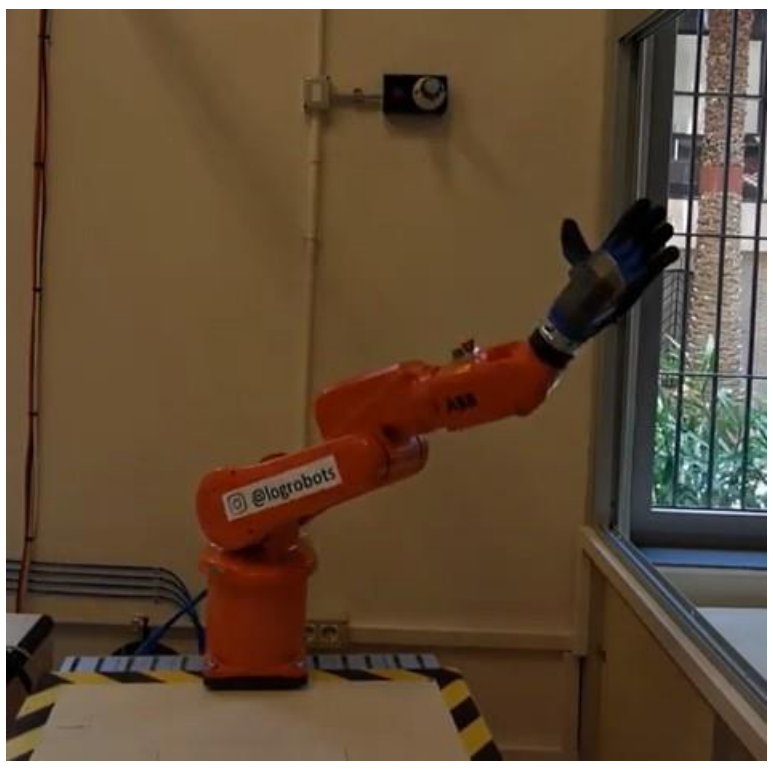

*Figure 11. Snapshot of test 1.*

The test was used to analyze the capacity of humans to distinguish between a human- like and a robot-like movement. For this, eight videos were recorded, with the robot performing four movements with a constant speed profile and four with a lognormal one. The following instructions were given to participants: “You will now be shown 8 short video sequences with actions carried out by a robotic arm. We are investigating new methods to make robot movements more human-friendly”.

The test had two parts:

- Part 1: Three videos with the same movement were performed with a uniform and lognormal speed profile. For each video, the participant was asked which of them was friendlier.

- Part 2: Five videos with just one movement repeated twice. For each video, the participant was asked to rate the friendliness of the movement. The scale was (0–10), where 0 was not at all friendly and 10 was very friendly.

The number of participants was 1088. Among these, 24% had knowledge of robotics, 30% did not, 22% had no technical knowledge and the remaining 24% did not comment. Women made up 44% and men 56%. The results did not show significant differences in terms of the knowledge of robotics or for the sex.
The results of the first part of the test showed that if the lognormal movement was shown last, the lognormal movement was considered friendlier than the uniform movement by 60% of participants. However, if the uniform one was shown last, the lognormal movement was considered friendlier by 49%.

The results of the second part of the test showed a preference for the uniform movement, with participants giving a friendliness rating of 6.4 with a standard deviation of 2.32 for the lognormal movement and 7.5 with a standard deviation of 2.17 for the uniform one. Therefore, the results of the test did not show any significant preference for either movement.

*5.2. Test 2*

This test was also carried out using Google forms (https://forms.gle/bS68BtBPcheo4 HNf7, accessed on 1 October 2022). It was based on a point moving on a screen with human or robotic motion, as shown in Figure 12.

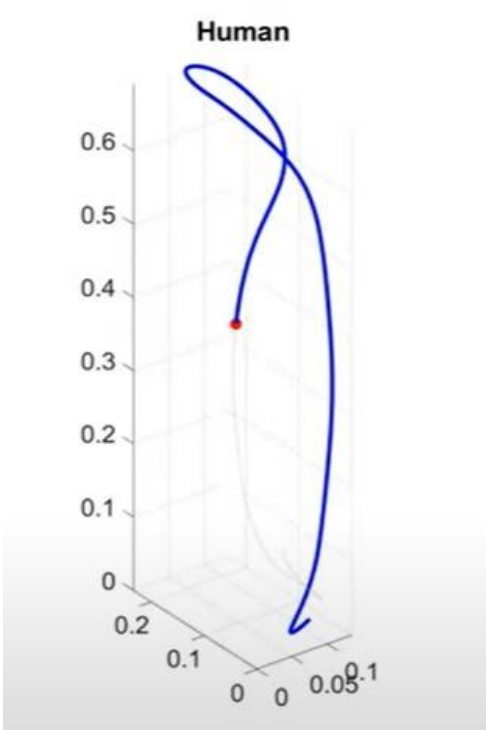

*Figure 12. Snapshot of test 2.*

The purpose of the test was to figure out if the subject could perceive any difference between robot and human movements. For this, several short videos of a red dot describing a trajectory in three dimensions were shown to the subjects. The path to be travelled was shown in light gray and the path travelled in blue. Participants were asked to indicate whether the observed movement was performed by a machine or a human.

The instructions given to participants were: "You will see a sequence of 17 short videos of movements made with a fingertip writing in the air. Some of them were carried out by a human, others by a machine. Please, indicate for each video who made the movement: A human or a machine".

The test had two parts:

- Part 1: Twelve videos were shown. The participants had to decide if the movement was made by a human or by a machine.
- Part 2: Five videos were shown with two executions of the same movement and each participant had to decide which movement was human or machine.

A total of 369 volunteers participated. Of these, 59% had computer skills. In terms of age distribution, 23% of those surveyed were under 16, 41% between 17 and 40 years old and the rest over 41. As in the first test, no significant differences were observed among the different groups analyzed.

In the first part of the test, the percentage of subjects who correctly chose human movement was 55% and the percentage who correctly chose robotic was 53%.

In the second part, the percentage of subjects who selected the correct movement was 53%.

Therefore, with this type of test, the subjects were not able to distinguish between robotic and human movements.

*5.3. Test 3*

This test was carried out using Google forms (https://docs.google.com/forms/d/e/1FAIpQLSehIW8wbO8tY-npOeVrDwbDNyT_2B5MPUFr19mpjgvyBew6fg/viewform, accessed on 1 October 2022). It was performed with a Universal Robots robot model UR3 at the Bialystok University of Technology, as shown in Figure 13. The steps for programming the speed profiles can be found at [15,20].

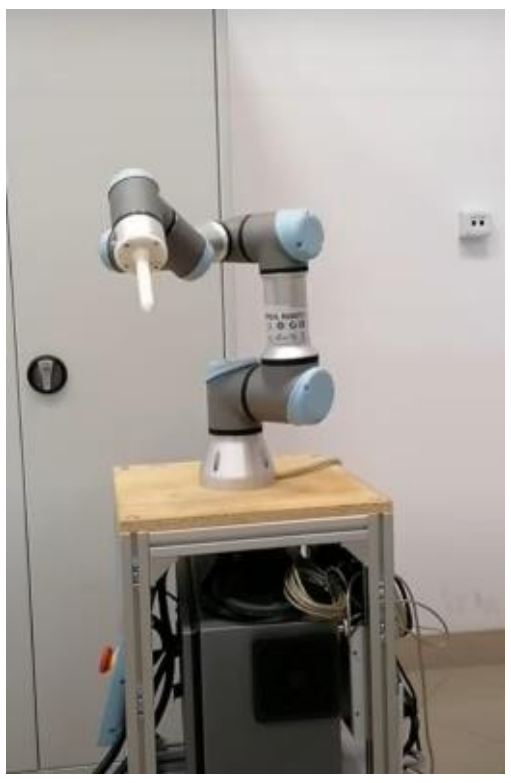

*Figure 13. Snapshot of test 3*

The purpose of this experiment was to analyze whether humans prefer human or robotic movement when looking at a robot moving.

The instructions given to the participants were: "Imagine that you are going to collaborate with a robotic arm. You will see some videos of a robot performing several movements with different speed profiles. For each of the presented movements, please score the confidence which you would collaborate with the robot. The questionnaire will take approximately 5 min. Please score the confidence you would have cooperating with a robot moving like this: 1 less confident. 5 more confident".

This test was carried out by 61 participants, 36% of whom had no knowledge of robotics, 41% with medium knowledge and the rest with high knowledge.

The result obtained indicated that the participants valued the uniform movement, with an index of 3.7 and a standard deviation of 0.87, against the lognormal movement, with an index of 3.4 and a standard deviation of 0.82.

There was no significant preference between the movements.

*5.4. Test 4*

In this test, an ABB robot was used to verify the influence of the movement typeon collaborative tasks. Previous tests were based on showing participants videos and analyzing whether they were able to differentiate human from robotic movement.

The aim of this test was to have the robot perform movements at safe speeds with a uniform and robotic speed profile. The participants were asked to touch the end of the robot with their index finger and to follow its movements, as shown in Figure 14. After this, they were asked which movements felt more comfortable. The tests were completed on-site by the participants with the procedure being as follows:

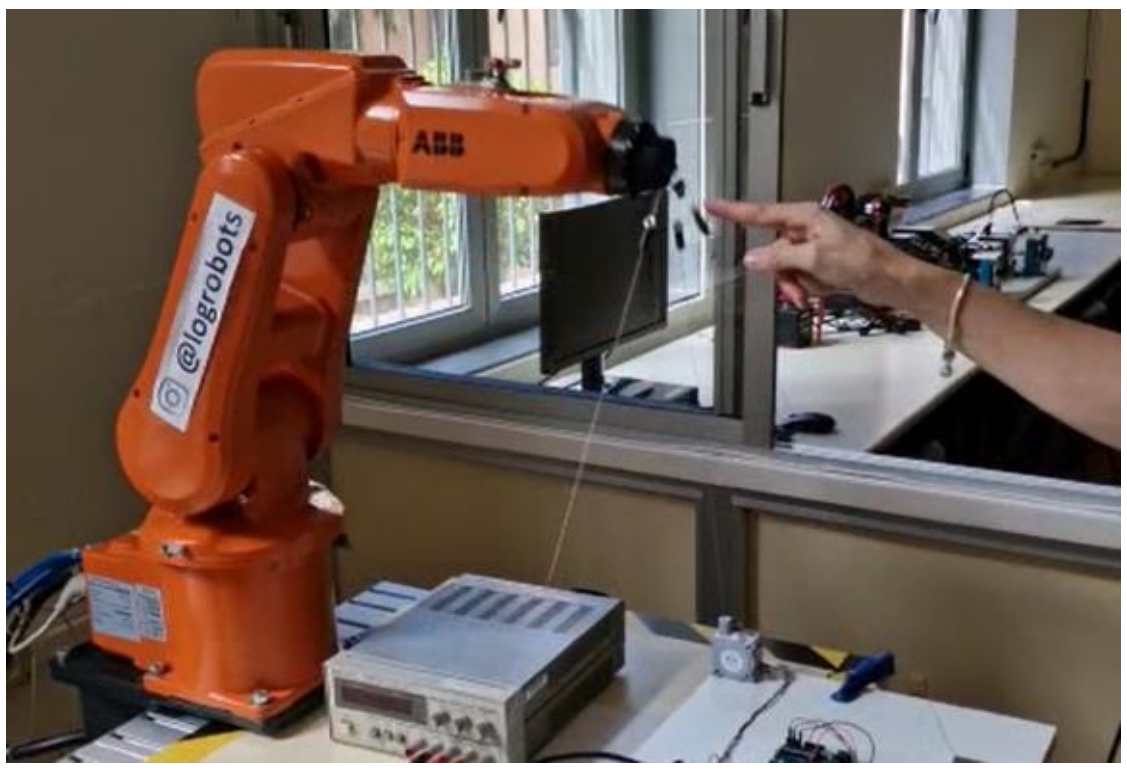

*Figure 14. Snapshot of test 4.*

The test has two parts:

- Part 1: The robot executed the Figure six times, the first three one way, followed bya 1-second pause and then three another way. The participants were asked to identify the one that was easiest to follow and also rate them. The rating was: 1 very unfriendly and 5 very friendly.
- Part 2: The same as above, but with a bigger figure.

The form given to the participants for each figure was:
The robot will perform a figure six times in a row, three with a speed profile (movement 1) and threewith the other speed profile (movement 2).

You must touch the robot's end with your finger and follow its movement. After that, answerthese questions.

Q1: Which of the robot movements did you find most friendly? Check the option.

- Movement 1.
- Movement 2.
- Both equal.

Q2: With a scale from 1 to 5, assess each of the previous movements, being 1 (very unfriendly) and5 (very friendly).

- Movement 1: _________
- Movement 2: _________

The figures drawn by the robot are those shown in Figure 7. The four on the left were shown randomly in the first part of the experiment and the one on the left was shown to all participants in the second part of the experiment. The choice of whether the first profile displayed was lognormal or at constant speed was also random.

This experiment was carried out by 118 participants, of whom 37% were women and 63% men, 31% had a low knowledge of robotics, 60% a medium knowledge and 9% a high knowledge. Regarding the age distribution, 53% were under 30, 13% between 31 and 50 and 51% were over 51. As in previous tests, no differences were observed related to the different groups.

The results obtained in the first part of the test showed that 42% of the participants preferred a lognormal movement, 44% preferred a uniform movement and 14% had an equal preference for both.
For the movements shown in second place, it was observed that if the last movement shown was the lognormal, the participants preferred it by 48% and trapezoidal by 36%. However, if the second shown was the trapezoidal, they preferred the lognormal by 36% and the trapezoidal by 54%. This behavior was observed in all tests in which participants had to choose between two options. Because the order of the movements was chosen randomly, this factor was not considered and only the global data were shown.

The results obtained in the second part of the test showed that 44% of the participants preferred the lognormal movement, 33% uniform and 22% preferred them equally.

The subjects were also asked to value each of the movements. A summary of the results, including the mean and the standard deviation of all ratings for the two figures and with both speed profiles, can be seen in Table 2.

*Table 2. Movement evaluation (1 unfriendly–5 very friendly).*

| | **Small Figures** | | **Big Figure** | |
|---|---|---|---|---|
| | **Lognormal** | **Trapezoidal** | **Lognormal** | **Trapezoidal** |
| Mean | 3.35 | 3.25 | 3.54 | 3.57 |
| Std. Dev | 1.10 | 1.16 | 1.21 | 1.18 |

**6. Discussion**

In the first test, the volunteers were shown robotic movements with a trapezoidal speed profile, the most common in robotic movements and a lognormal profile more like human movement. The humans' preference for the two movements was similar and it was observed that they preferred the last movement shown. Moreover, in terms of the friendliness of both movements, the movement at constant speed was slightly better valued. From this test, no appreciable preference for any of the movements can be deduced.

In the second test, writing movements performed by a machine and humans were shown on a screen to the subjects. There was no clear preference for the human movement over the other. However, in a similar experiment, this paper [21] showed that participants had a slight preference for human-like movements.

In the third test, participants watched videos of a robot with shorter movements than in test 1 and with both speed profiles. When assessing them, the valuation of movements at constant speed was slightly higher than that for lognormal movements.

Finally, in the robot interaction test, the participants also did not show any significant preference for either of the speed profiles. In fact, the ratings were very similar.

In all tests, it was observed that the participants do not show a clear preference for movement at constant speed as compared to a more human movement, both in terms of visualizing the movement and of relating to the robot.

Regarding the hypotheses raised in Section 5, the results of the tests seem to show that the robotic movement, if the movement at constant speed is taken as such, does not cause any rejection and is sometimes even preferred to the human one. Regarding the human–robot interaction, the volunteers were equally comfortable with both movements.

## 7. Conclusions

In this paper, we analyzed whether the movement of a robotic arm with a human speed profile enhances its relation and interaction with humans. To this end, the equivalence between the human and the robotic arm was analyzed, as was the difference between human and robotic movements. It was shown that the speed profile of a robotic movement has a constant speed, while a human one is bell-shaped.

Due to the unavailability of commands for the robot to generate human movements, such movements were synthesized using the lognormal model. A novel procedure to implement these movements in a robotic arm was shown and verified.

Four tests were carried out. The first three showed that humans do not have a clear preference for the human movement or for the robotic when they look at the movement. The fourth showed the same result when a human interacts with the robot.

The results of the tests seem to show that robotic movement does not cause any rejec- tion and is sometimes even preferred by humans. Regarding the human–robot interaction, the volunteers were equally comfortable with both movements.

This paper opens the door to new tests: on the one hand, to check if factors such as speed and the amplitude of movement influence the preference for one movement or another and on the other, to analyze other types of movements [22].